\documentclass[runningheads]{llncs}

\usepackage{eccv}

\usepackage{eccvabbrv}

\usepackage{graphicx}
\usepackage{booktabs}
\usepackage{amsmath}
\usepackage{amssymb}
\usepackage{enumitem}
\usepackage{multirow}
\usepackage{xcolor}
\usepackage{subcaption}
\usepackage{bbding}
\usepackage[accsupp]{axessibility}  

\usepackage{hyperref}

\usepackage{orcidlink}

\begin{document}

\title{InterPet4D: A Multimodal 4D Human-Pet Interaction Dataset for Pet Motion Generation}

\titlerunning{InterPet4D}

\author{Yichen Peng*\inst{1} \and
Jyun-Ting Song*\inst{1,2} \and
Chen-Chieh Liao*\inst{1} \and \\
Kris Kitani\inst{2} \and
Hideki Koike\inst{1} \and
Erwin Wu\inst{1}
}

\authorrunning{Y.~Peng et al.}

\institute{Institute of Science Tokyo \and
Carnegie Mellon University
}
\maketitle

\begin{figure*}[tbhp]
    \centering
    \includegraphics[width=0.88\textwidth]{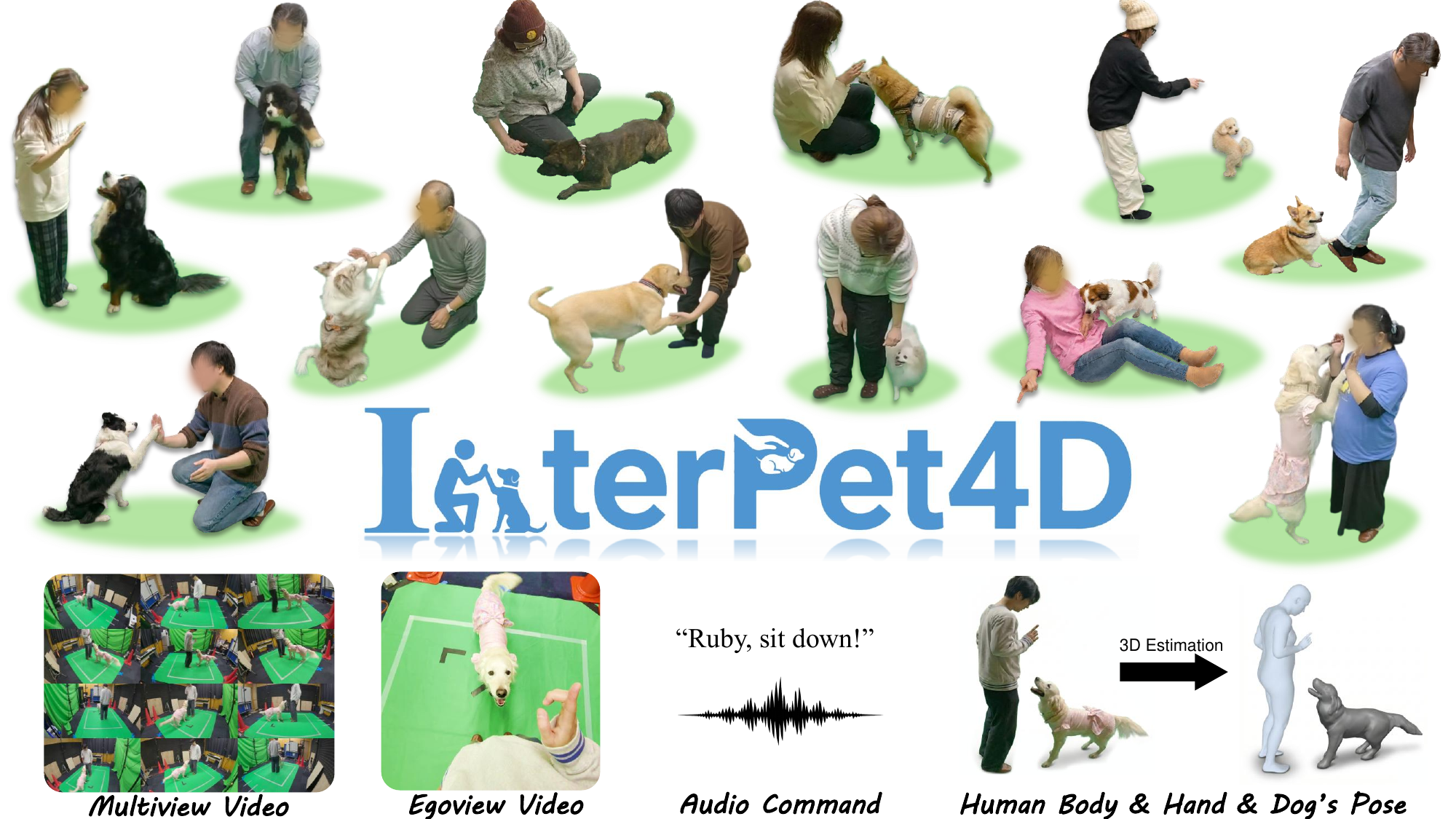}
    \caption{The \textbf{InterPet4D} multimodal Human-Pet Interaction Dataset.}
    \label{fig:teaser} \vspace{-34pt}
\end{figure*}

\begin{abstract}

Human-pet interaction estimation and generation remain underexplored due to the absence of high-quality large-scale dataset. We present \textbf{InterPet4D}, the first multimodal dataset capturing natural interactions between humans and dogs. Using a synchronized multi-view capture system, we record human–dog obedience tasks and provide annotations for both humans and dogs, including multiview and egocentric videos, segmentations, 2D/3D keypoints, meshes, and audio tracks. \textbf{Interpet4D} consists of 6.8 million frames collected from 13 dogs of 11 breeds interacting with 23 human participants. We further introduce the \textbf{InterPetMoGen} framework for human-pet interaction motion generation. Our proposed model achieves an FID score of 11.21, substantially outperforms the Seq2Seq or DiT baselines, demonstrating the effectiveness of Interpet4D for modeling realistic human–pet interactions.
\end{abstract}
\noindent{\footnotesize * Equal contribution.}

\section{Introduction}
\label{sec:introduction}

Human–animal interactions exhibit temporally coordinated and mutually responsive movement patterns, where the motion of one agent directly influences and adapts to the other. Modeling such structured relationships is fundamental for understanding cross-species behavior and has important applications in socially aware robotics, virtual agents, animation, and behavioral analysis.

However, most existing research on interactive behavior has focused on human–human or human–object interactions, largely due to the absence of large-scale and realistic human-animal datasets. Capturing such interaction data presents unique challenges, as it requires structured and repeatable coordination between human and animal participants, which is difficult to achieve without controlled training, making large-scale data collection particularly challenging. Furthermore, close-range human–animal interactions can result in severe cross-occlusion, which degrades the reconstruction of fine-grained details on the participants.

To address these challenges, we introduce \textbf{InterPet4D}, the first large-scale multimodal 4D dataset of naturalistic human-dog interactions. Our dataset captures synchronized multiview, egocentric, audio, and 3D motion data from 23 participants and 13 dogs. We design a systematic interaction protocol covering 4 categories of common interactions: \emph{petting}, \emph{commanding}, \emph{calling}, and \emph{free-form}, enabling structured analysis of dog behavior. Figure~\ref{fig:teaser} provides an overview of the InterPet4D dataset, including multiview video, egocentric video, audio commands, and reconstructed human and dog motion. To facilitate future research, we release InterPet4D on Hugging Face:
\url{https://huggingface.co/datasets/ohicarip/interpet4d}.

Beyond the dataset, we propose \textbf{InterPetMoGen (IPMG)}, a Motion GPT-based framework for gesture-to-pet motion generation. Given a sequence of human hand/body gestures and accompanying audio, \textbf{IPMG} generates plausible 3D dog motion responses conditioned on human gestures and audio.
The model adopts a MotionGPT-style autoregressive transformer that predicts discrete pet motion tokens learned by a PetVAE tokenizer.
We further introduce modality-aware attention (MAA) masks that enable coarse-to-fine motion generation by combining bidirectional conditioning with causal autoregressive decoding.

Our contributions are summarized as follows:
\begin{itemize}[leftmargin=15pt,itemsep=2pt]
    \item We introduce \textbf{InterPet4D}, the first large-scale multimodal 4D dataset of human–pet interactions, containing 6.8M synchronized frames with multi-view RGB video, egocentric video, audio, and reconstructed 3D motion of both humans and dogs across 23 participants and 13 dogs of 11 breeds.
    \item We design a systematic interaction protocol covering 4 categories of human–dog interactions with standardized annotation pipelines, enabling structured analysis of cross-species behavior.
    \item We propose \textbf{IPMG}, a MotionGPT-based framework including a PetVAE tokenizer and MAA module that generates diverse and realistic dog motion responses conditioned on human gestures and audio signals.
\end{itemize}

\section{Related Work}
\label{sec:related}



\subsection{Human-centered Interaction Datasets}

The study of human interactions with the physical world has progressed rapidly.
GRAB~\cite{taheri2020grab} captures whole-body grasping of objects with detailed hand motion.
BEHAVE~\cite{bhatnagar2022behave} provides multi-view recordings of humans interacting with rigid objects.
ARCTIC~\cite{fan2023arctic} focuses on articulated object manipulation with dexterous hand motion.
For human--human interaction, InterHuman~\cite{liang2024intergen} proposes a large-scale dataset and a diffusion-based model for two-person motion generation.
BUDDI~\cite{muller2024buddi} reconstructs 3D human--human close interactions from monocular images.
These works demonstrate that modeling interactive dynamics requires capturing the joint distribution of all interacting agents.
Our work extends this principle to the human-pet domain, where the morphological asymmetry between human and animal introduces additional challenges.

\subsection{Animal Pose and Shape Estimation}

Estimating the 3D pose and shape of animals has attracted growing attention.
SMAL~\cite{zuffi20173d} introduces a parametric 3D body model for animals learned from toy figurines.
Subsequent works focus on dogs, including WLDO~\cite{biggs2020wldo}, BARC~\cite{rueegg2022barc}, and BITE~\cite{rueegg2023bite}, which improve monocular 3D reconstruction using breed priors and contact constraints, as well as DogMo~\cite{wang2025dogmolargescalemultiviewrgbd}, which reconstructs dog motion from monocular videos,  and AnimalAvatar~\cite{AnimalAvatars2024}, which reconstructs animatable 3D animals and motion from videos.
RigAnything~\cite{liu2025riganything} further enables automatic skeletal rigging for arbitrary animal meshes. For pose estimation, DeepLabCut~\cite{mathis2018deeplabcut} provides a widely-used markerless framework across species, while RatBodyFormer~\cite{higami2025ratbodyformerratbodysurface} applies transformer models to estimate rodent body pose.
On the data side, Animal Kingdom~\cite{ng2022animal} and COP3D~\cite{sinha2023common} provide large-scale animal video datasets for recognition and 3D reconstruction.
However, existing datasets mainly focus on single-animal settings and rarely capture human–animal interactions or multimodal signals such as audio and human motion.

\subsection{Conditional Motion Generation}

Generating human motion conditioned on various signals has seen remarkable progress.
Action-conditioned methods such as ACTOR~\cite{petrovich2021action} use VAE-based architectures for class-conditional generation.
Text-conditioned approaches have flourished with MDM~\cite{tevet2023human}, which applies diffusion models to motion generation from text descriptions, and T2M-GPT~\cite{zhang2023t2m}, Duolando~\cite{siyao2024duolandofollowergptoffpolicy}, MDLS~\cite{chen2023executing}, which uses VQ-VAE with GPT-based autoregressive generation.
MotionDiffuse~\cite{zhang2022motiondiffuse}, FloodDiffusion~\cite{cai2026flooddiffusiontailoreddiffusionforcing} enable fine-grained text-driven motion synthesis.
MoMask~\cite{guo2024momask} introduces masked modeling for efficient motion generation.
In the audio domain, co-speech gesture generation methods~\cite{liu2022beat,yi2023generating,peng2026dyaditmultimodaldiffusiontransformer,Liu2024emage} synthesize body and hand gestures from speech audio.
MotionGPT~\cite{jiang2024motiongpt} treats motion as a language and unifies multiple motion tasks in a single framework.
However, all existing methods operate within a single species while our work pioneers the cross-species setting, generating animal motion conditioned on human multi-modal signals.

\section{InterPet4D Dataset}
\label{sec:dataset}
\begin{figure}[tbhp]
\vspace{-5pt}
    \centering
    \includegraphics[width=0.95\linewidth]{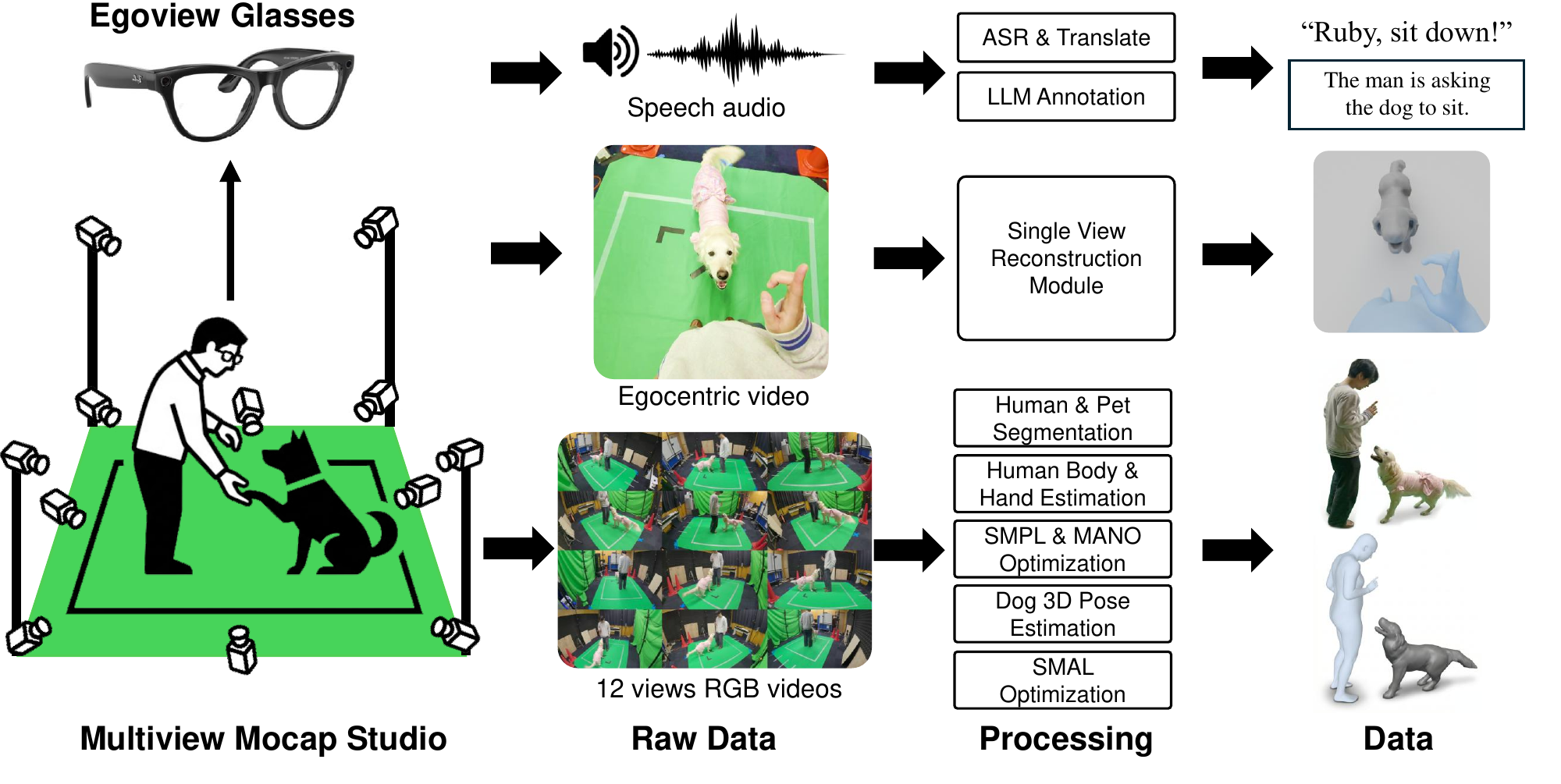}
    \vspace{-10pt}
    \caption{\textbf{Data collection setup.} Our capture environment consists of 12 synchronized third-person RGB cameras arranged in a square, complemented by Rayban Meta Glasses worn by the participant providing an egocentric view and audio.} 
    \label{fig:setup}
\end{figure}

\subsection{Data Collection Setup}
\label{sec:collection}
We establish a multi-sensor capture environment to record synchronized multimodal data of human-pet interactions. The capture space is a 6m$\times$6m area.

\noindent\textbf{Multi-view cameras studio.}
We deploy 12 wire-synchronized GoPro Hero13 cameras (1920$\times$1080, 60\,fps) arranged in a square configuration around the capture area, providing comprehensive multi-view coverage of both the human and the dog, as shown in Figure~\ref{fig:setup}. Different from traditional human-target mocap studio, 8 out of 12 cameras are placed in the lower-ground area to achieve a better view of the pet and human's hand.

\noindent\textbf{Egocentric view and audio.}
In addition to the third-person cameras, each participant wears a pair of Ray-Ban 2 Meta glasses, which provide an egocentric RGB video and audio stream capturing the first-person perspective of the interaction.
The videos are synchronized with the GoPro cameras through software-based alignment using a hand-clapping signal, resulting in a maximum synchronization error of one frame.
This egocentric view is particularly valuable for capturing close-range hand-pet interactions that can't be recorded from the 3rd-person-views.

\noindent\textbf{Participants.}
To ensure stable interactions, we collaborated with a local pet training agency and recruited 11 experienced trainers together with different breeds of trained dogs. 
In addition, 12 volunteers participate in the recordings, resulting in a total of 23 human subjects. Two untrained puppies were also involved for diversity, which results in 13 dogs spanning 11 breeds, covering a wide range of body sizes, as shown in Table~\ref{tab:stats}. All adult dogs were trained in basic obedience tasks such as sitting, turning, and calling gestures, enabling consistent execution of the interaction protocol and still allowing natural behavior.

\begin{table}[t]
\centering
\caption{\textbf{InterPet4D dataset statistics.} Our dataset provides synchronized multimodal 4D data of human--dog interactions across diverse participants, breeds, and interaction types.}
\begin{minipage}{0.625\linewidth}
\centering
\setlength{\tabcolsep}{8pt}
\begin{tabular}{ll}
\toprule
\textbf{Data Attribute} & \textbf{Value} \\
\midrule
No. of dogs & 13 \\
No. of dog breeds & 11 (+2 puppies) \\
No. of human & 23 (incl. 11 trainers) \\
Third-person cameras & 12 (1920$\times$1080, 60 fps) \\
Egocentric camera & 1 (1200$\times$1600, 60 fps) \\
Total recording & 161 sessions \\
Total frames & 6.83M \\
Human representation & Pose, SMPL-X, MANO \\
Dog representation & Pose, SMAL \\
Audio & Aligned voice command \\
Text Caption & Annotated each clip \\
\bottomrule
\end{tabular}
\end{minipage}
\hfill
\begin{minipage}{0.365\linewidth}
\centering
\includegraphics[width=1.03\linewidth]{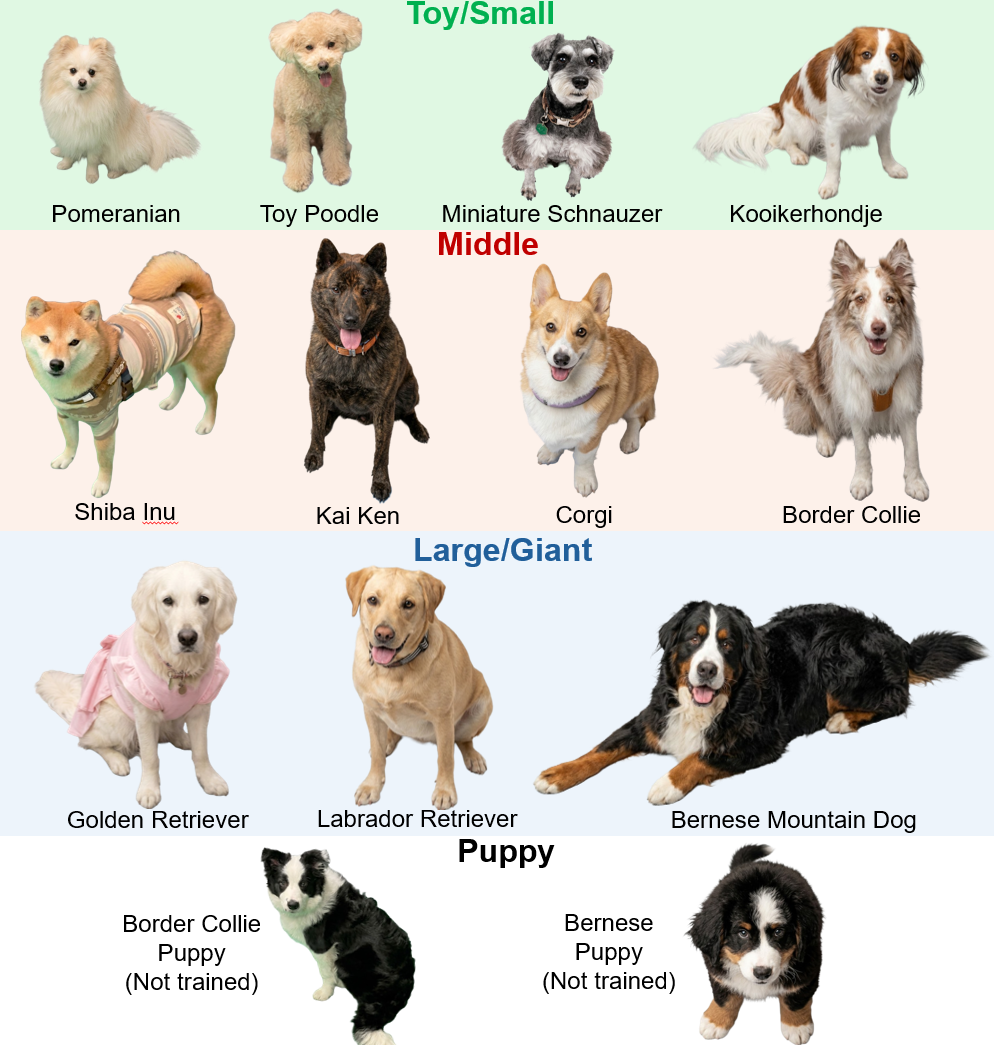}
\end{minipage}

\label{tab:stats}

\end{table}

\begin{table}[t]
\centering
\caption{\textbf{Comparison with other Interaction or Pet datasets.} InterPet4D is the first dataset providing synchronized human and pet 3D interactive motion with audio.}
\label{tab:comparison}
\resizebox{\linewidth}{!}{
\begin{tabular}{lcccccccc}
\toprule
\textbf{Dataset} & \textbf{Subjects} & \textbf{No.Human} & \textbf{No.Pet} & \textbf{Audio} & \textbf{Interactive} & \textbf{Views} & \textbf{Frames} & \textbf{Public}  \\
\midrule
InterHuman~\cite{liang2024intergen} & Human & 12 pairs & -- & -- & \Checkmark & 76 & 107M &\Checkmark \\
Seamless Inter.~\cite{agrawal2025seamlessinteractiondyadicaudiovisual} & Human & 4000+ & -- & \Checkmark & Speech-only & 1 & 400M+ &\Checkmark \\
\midrule
CoP3D~\cite{sinha2023common} & Dog \& Cat & -- & 4200 & -- & -- &  1 & 0.6M &\Checkmark  \\

DogMo~\cite{wang2025dogmolargescalemultiviewrgbd} & Dog & -- & 10 & -- & -- &  5 & 1M &   -- \\
\midrule
\textbf{InterPet4D} & Human+Dog & 23 & 13 & \Checkmark & \Checkmark  & 12+Ego & 6.8M &  \Checkmark *\\
\bottomrule
\multicolumn{9}{r}{
\scriptsize{*The full dataset including annotations will be released upon publication.}}
\end{tabular}
} 
\vspace{-20pt}
\end{table}

\subsection{Interaction Taxonomy}
\label{sec:taxonomy}

We design a structured interaction protocol covering four categories of common human-dog interactions, each emphasizing different communication modalities:

\begin{enumerate}[leftmargin=15pt,itemsep=0pt]
    \item \textbf{Petting} -- The participant physically touches or strokes the dog, involving close-range hand motion and the dog's postural response (\eg, leaning in, tail wagging).
    \item \textbf{Commanding} -- The participant issues verbal and gestural commands (\eg, ``sit'', ``stay'', ``shake'', ``turn around''), and the dog responds with trained behaviors.
    \item \textbf{Calling} -- The participant calls the dog from a distance using voice and hand beckoning repetitively, capturing the dog's locomotion and attention shifts.
    \item \textbf{Free-form} -- Unscripted interactions allowing participants to interact naturally, capturing the full diversity of everyday human--dog dynamics, including fetch, tug-of-war, and chase. 
\end{enumerate}

Each participant--dog pair performs 3-4 sessions per category, with each session lasting about 60 seconds.

\subsection{Dataset Statistics}
\label{sec:statistics}
InterPet4D is the first dataset to provide multimodal recordings of human–pet interactions with synchronized 3D motion for both agents. Table~\ref{tab:stats} summarizes the key statistics of InterPet4D. To enable fair and reproducible benchmarking, we provide a fixed 80:20 train/validation split. All dogs appear in both sets, resulting in 200 training clips and 40 validation clips. Table~\ref{tab:comparison} compares InterPet4D with other human-interaction and pet datasets. For more details, such as dataset analysis and examples, please check the supplementary documents.

\section{Data Processing Pipeline}
\label{sec:annotation}

\subsection{Human Reconstruction}
We represent humans using the SMPL~\cite{loper2015smpl}, SMPL-X~\cite{pavlakos2019expressive}, and MANO~\cite{romero2022embodied} parametric models. Human reconstruction is decomposed into body and hand reconstruction.

\noindent \textbf{Body Reconstruction}
The human body pose is first estimated from multi-view third-person cameras and fitted to the SMPL~\cite{loper2015smpl} model. We first localize the human in each view using Mask R-CNN~\cite{he2017mask} with DeepSORT tracking~\cite{wojke2017simple}, followed by manual filtering to remove failure cases. We then estimate 2D body keypoints using HRNet-WholeBody~\cite{jin2020whole} for each detected bounding box and triangulate the multi-view 2D keypoints to obtain 3D joint estimates. To further refine the 3D body pose, we minimize an energy function that incorporates body symmetry, temporal smoothness, and temporal bone-length constraints~\cite{khirodkar2024harmony4d}. Finally, an SMPL~\cite{loper2015smpl} body model is fitted to the refined 3D joint estimates to recover the full-body pose and mesh for each frame.

\noindent \textbf{Hand Reconstruction}
Hand pose is estimated using a pipeline similar to body reconstruction. We first localize the hands in each view by running YOLO~\cite{hussain2023yolo} and comparing the detections with the reprojected SMPL hand mesh. Using the resulting close-up hand crops, we apply WiLoR~\cite{potamias2025wilor} to estimate 2D hand keypoints in each view. Following the same scheme as body reconstruction, the multi-view 2D hand keypoints are triangulated to obtain initial 3D estimates and then refined using the same energy formulation as in body reconstruction, augmented with a visibility-aware optimization~\cite{songcontact4d, shinbodycontact4d}. Finally, we fit MANO~\cite{romero2022embodied} to the refined 3D hand keypoints to recover hand pose, and fit SMPL-X~\cite{pavlakos2019expressive} to the combined body and hand keypoints to recover the full-body.


\subsection{Dog Reconstruction}

We represent dogs using sparse 3D keypoints and the SMAL~\cite{zuffi20173d} body model. For each view, we first detect the dog using RTMDet~\cite{lyu2022rtmdet} and estimate 2D dog keypoints using HRNet-W32~\cite{cheng2020higherhrnet}. We then run Mask R-CNN~\cite{he2017mask} to segment humans, and use the resulting human masks to filter out 2D dog keypoints that fall in human-occluded regions. The filtered 2D keypoints from multiple views are triangulated to obtain per-frame 3D joint estimates, augmented with an RTS smoother~\cite{sarkka2008unscented} to improve temporal consistency. Finally, we fit the SMAL~\cite{zuffi20173d} model to the reconstructed 3D keypoints as SMALify~\cite{biggs2018creatures}. Since mesh fitting from sparse keypoints is highly underconstrained, we predefined the shape parameters by manually fitting SMAL~\cite{zuffi20173d} to a single representative frame in a canonical pose to stabilize the reconstruction. 

\subsection{Audio and Text}

\noindent\textbf{Audio Extraction \& Translation}
Raw audio is captured by the built-in microphones of the Ray-Ban Meta glasses. Among the 13 dogs, 11 are trained with Japanese commands, one with English commands, and one with Mandarin Chinese commands. To unify the language modality, we use an automatic speech recognition (ASR) system~\cite{Qwen3ASR} to transcribe the spoken commands and translate them into English. In addition, the temporal boundaries of each command are automatically annotated using Qwen3-ForcedAligner~\cite{Qwen3ASR}.

\noindent\textbf{Interaction labels.}
Each sequence is annotated with an interaction category (Section~\ref{sec:taxonomy}) and a textual description of the human–pet behavior (e.g., ``The man first calls the dog, then commands the dog to sit and turn around.''). 
The textual annotations are automatically generated from the English transcripts of the recorded audio using a Qwen3-based large language model~\cite{Qwen3ASR} and only include descriptions of the verbal actions.


\begin{figure}[t]
    \centering
    \includegraphics[width=0.95\linewidth]{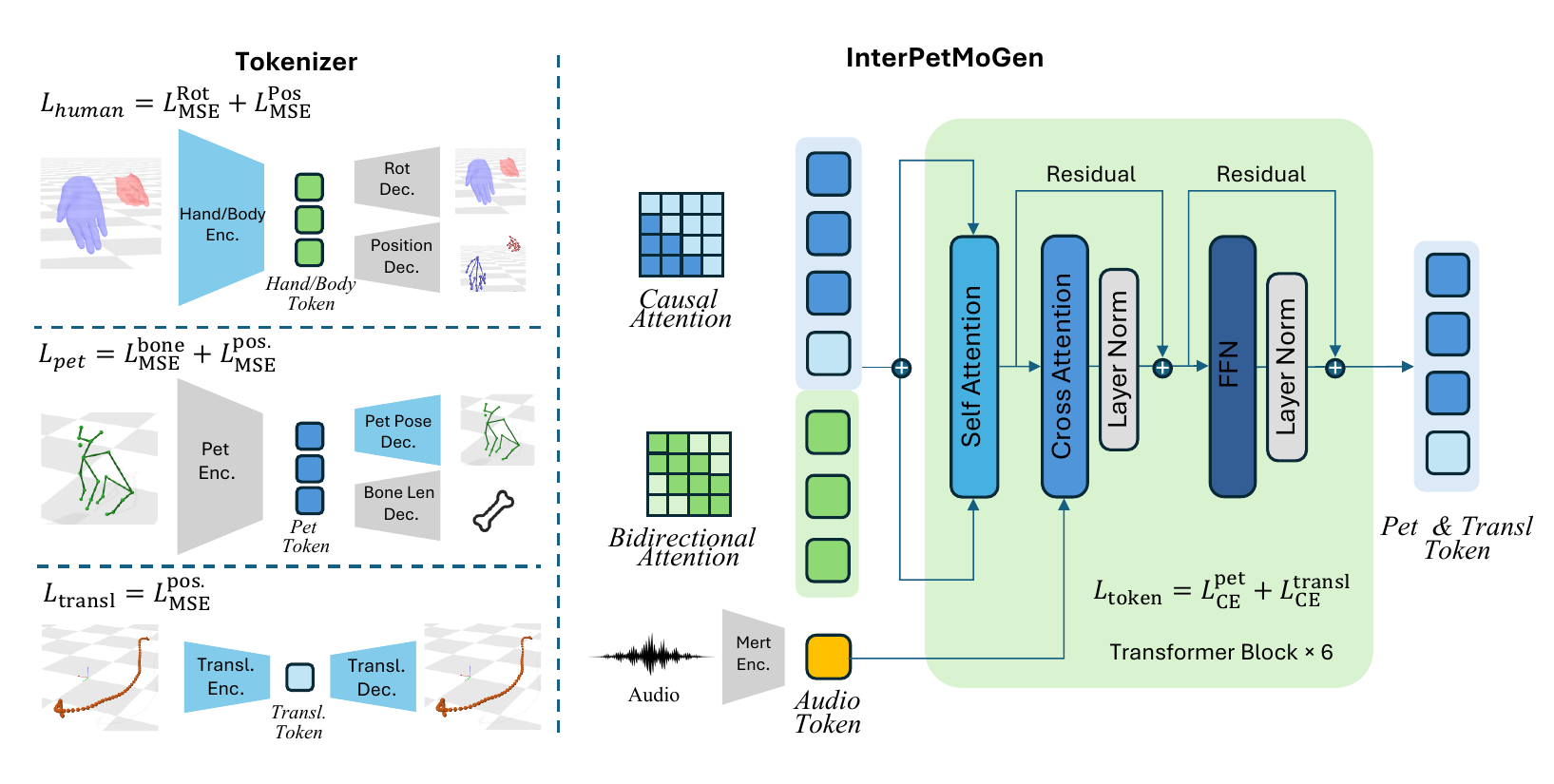}
    \caption{\textbf{InterPetMoGen architecture.} Human gesture tokens and audio features are encoded by modality-specific encoders and fused through an autoregressive transformer. The model generates global translation and pet motion tokens conditioned on human body, hand, and audio tokens. The generated tokens are decoded into continuous dog motion using the PetVAE decoder.}
   
    \label{fig:framework}
\end{figure}

\section{InterPetMoGen: Human-to-Pet Motion Generation}
\label{sec:method}

\subsection{Problem Formulation}
\label{sec:formulation}
Given a human motion sequence 
$\mathbf{H} = \{h_t\}_{t=1}^{T}$, 
where $h_t$ contains the 3D joint positions and rotations of $J_b$ body joints and $J_h$ hand joints at frame $t$, 
and a corresponding audio feature sequence 
$\mathbf{A} = \{a_t\}_{t=1}^{T}$, 
where $a_t \in \mathbb{R}^{d_a}$ denotes the audio feature at frame $t$, 
Our goal is to generate a pet motion sequence 
$\mathbf{P} = \{p_t\}_{t=1}^{T}$, 
where $p_t \in \mathbb{R}^{J_p \times 3}$ represents the 3D positions of $J_p$ pet joints.

\noindent To better capture the distinct characteristics of human body and hand movements and learn their relationship towards pet reaction, we propose \textbf{InterPetMoGen}.
We represent body and hand as two separate streams and tokenize them independently in our model. The mapping from human motion and audio to pet motion is inherently one-to-many: the same human input may correspond to multiple plausible pet responses. 
Therefore, we aim to learn the conditional distribution $p(\mathbf{P} \mid \mathbf{H}, \mathbf{A})$ rather than a deterministic mapping.

\subsection{Model Overview}
As illustrated in Figure~\ref{fig:framework}, \textbf{IPMG} consists of two major components. \textbf{Multi-modality tokenization}, which converts continuous signals from different modalities into discrete tokens. \textbf{Autoregressive transformer generation}, which models the conditional distribution of pet motion tokens given human motion and audio tokens.

\noindent Specifically, human body and hand gestures are first encoded into motion tokens using modality-specific encoders. Pet motion is represented using a discrete latent space learned by a PetVAE. In addition, global translation and audio features are converted into dedicated tokens. All tokens are then concatenated and fed into a transformer model, which predicts the next pet motion token conditioned on the multi-modal context.
The generated pet tokens are finally decoded back to continuous motion sequences using the PetVAE decoder.

\subsection{Multimodal Condition Encoder}
\label{sec:encoder}

\textbf{Pet Motion Tokenizer (PetVAE).} Pet motion exhibits different kinematic structures from human motion. To obtain a compact yet expressive representation, we learn a discrete latent space using a PetVAE.

\noindent Given a pet motion sequence $\mathbf{P}$, the encoder maps continuous joint positions into latent embeddings, which are then quantized using a learnable codebook to obtain discrete pet tokens. The decoder reconstructs the full pose sequence from the quantized embeddings. In addition to reconstructing joint positions, we utilize a \textbf{Bone Length Regressor} to enforce skeletal consistency. Specifically, alongside the pose decoder that predicts 3D joint positions, an auxiliary regressor predicts the bone length of each skeletal segment. This auxiliary task encourages the model to preserve the underlying skeletal structure when reconstructing poses.
To obtain a compact discrete representation of pet motion, we train a PetVAE to encode pose sequences into a codebook of motion tokens.
The model is optimized with the reconstruction objective
$\mathcal{L}_{pet} = \mathcal{L}^{pos}_{MSE} + \mathcal{L}^{bone}_{MSE} + \mathcal{L}_{VQ} + \beta \mathcal{L}_{commit}$,
where $\mathcal{L}^{pos}_{MSE}$ supervises joint position reconstruction and $\mathcal{L}^{bone}_{MSE}$ enforces bone-length consistency.
$\mathcal{L}_{VQ}$ and $\mathcal{L}_{commit}$ follow the standard VQ-VAE formulation~\cite{oord2018neuraldiscreterepresentationlearning}.
This objective encourages the learned motion tokens to preserve both accurate pose geometry and realistic skeletal proportions.

\noindent \textbf{Human Motion Tokenizer (Hand + Body).} Human motion contains both body movements and detailed hand gestures. To better capture their distinct motion patterns, we tokenize body and hand joints as two separate streams.

\noindent Given the human motion sequence $\mathbf{H} = \{\theta^{body}, \theta^{hand}, \mathbf{J}^{body}, \mathbf{J}^{hand}\}$ derived from SMPL body parameters and MANO hand parameters, an encoder maps the continuous joint rotations and positions into latent embeddings, which are then quantized using the codebook to obtain human motion tokens. Two decoders reconstruct joint rotations and positions from the quantized embeddings. It is trained with the following objective $\mathcal{L}_{human} = \mathcal{L}^{pos}_{MSE} + \mathcal{L}^{rot}_{MSE} + \mathcal{L}_{VQ} + \beta \mathcal{L}_{commit}$, where $\mathcal{L}^{rot}_{MSE}$ and $\mathcal{L}^{pos}_{MSE}$ supervise the reconstruction of joint rotations and positions, respectively.

\noindent \textbf{Translation Tokenizer.} 
Global root translations are modeled separately using a lightweight translation encoder–decoder module.
The encoder maps the root translation sequence $\mathbf{T}$ into latent embeddings, which are then quantized into translation tokens. By decoupling global trajectory from local joint motion, the translation tokens capture coarse motion dynamics and serve as an intermediate representation between human conditioning tokens and fine-grained pet pose tokens in the autoregressive generation process.

\noindent \textbf{Audio Tokenizer.}
We extract audio features using the pretrained MERT~\cite{li2023mert} and project them into token embeddings.
These tokens provide temporal cues for modeling human–pet interactions.

\subsection{Autoregressive Transformer}
\label{sec:transformer}
Given the multi-modal token sequence, we train a transformer to model the conditional distribution of pet motion tokens.

\noindent Human motion tokens serve as conditioning signals, while translation and pet motion tokens are generated autoregressively. Each transformer block contains self-attention, cross-attention, and
feed-forward layers.

\noindent \textbf{Modality-Aware Attention (MAA).} Human–pet interaction exhibits a natural hierarchical structure, where human motion provides global context, translation captures coarse motion, and pet pose represents fine-grained movements. To reflect this dependency, we design a modality-aware attention mechanism that controls information flow between tokens.

\noindent Let the multi-modal token sequence be $\mathbf{T} = [\mathbf{T}_{human}, \mathbf{T}_{transl}, \mathbf{T}_{pet}]$, where $\mathbf{T}_{human}=[\mathbf{T}_{hand}, \mathbf{T}_{body}]$ denotes human motion tokens, and $\mathbf{T}_{transl}$ and $\mathbf{T}_{pet}$ denote translation and pet motion tokens. Human tokens use bidirectional attention as global conditioning, while translation and pet tokens attend to preceding tokens only, forming a coarse-to-fine generation hierarchy.

\noindent \textbf{Audio Cross-Attention.}
Audio tokens are incorporated through a cross-attention module, providing temporal cues for interaction timing. The model predicts the next token with the objective $\mathcal{L}_{token} = \mathcal{L}^{pet}_{CE} + \mathcal{L}^{transl}_{CE}.$

\noindent During inference, tokens are generated autoregressively. Pet motion tokens are decoded by the PetVAE decoder, while translation tokens are decoded to recover global motion. The final pet motion is obtained by combining decoded poses with the predicted translations.


\section{Experiments}
\label{sec:experiments}
\subsection{Implementation Details}
\label{sec:implementation}

\noindent All models are implemented in PyTorch and trained on a single NVIDIA H100 GPU. We downsample the raw dataset to 30\,fps and split per-dog into 80\%/20\% train/val sets. All motion sequences are segmented into clips of length $T=300$ frames (approximately 10 seconds), with zero-padding applied to shorter sequences. During training, we apply a sliding window with a 50-frame stride to increase sample diversity. For IPMG training, these sequences are further chunked into 10-second windows, resulting in approximately 87.5 minutes of motion data. All sequences are normalized by subtracting the waist position in the first frame and aligning them to a canonical coordinate system based on the initial facing direction.

\noindent IPMG is trained in two stages. We first train four VQ-VAEs to tokenize each modality using a shared codebook size of 512 and temporal downsampling. We then train a 28M-parameter prefix-LM GPT to autoregressively generate dog motion tokens conditioned on motion tokens, with MERT audio features injected via cross-attention. The VQ-VAEs and GPT are trained for 500 and 300 epochs, respectively. More details are provided in the appendix.

\subsection{Evaluation Metrics}
We evaluate the generated motions using commonly used metrics in motion generation, including Fréchet Inception Distance (FID), Retrieval Precision (R-Precision), and Diversity (Div).

\noindent \textbf{Fréchet Inception Distance (FID)$\downarrow$}
FID measures the distributional similarity between generated motions and ground-truth motions. Lower values indicate that the generated motions are closer to real motion data. We report FID in both kinetic ($FID_k$) and static ($FID_s$) feature spaces.

\noindent \textbf{Retrieval Precision (R-Precision)$\uparrow$}
R-Precision evaluates how well the generated pet motions align with the conditioning signals. We report R-Precision with respect to hand ($R_{Prec.}^{hand}$) and body ($R_{Prec.}^{body}$) conditioning signals.

\noindent \textbf{Diversity (Div)$\uparrow$}
Diversity measures the variability of generated motion samples. Higher values indicate that the model produces more diverse motion patterns. We report diversity in both kinetic ($Div_k$) and static ($Div_s$) feature spaces.

\subsection{Baselines}
\label{sec:baselines}
\noindent Since no prior methods exist for human-to-pet gesture-to-motion generation, we compare our method with two representative architectures for motion generation:


\noindent \textbf{Seq2Seq-Transformer} adopts a standard encoder-decoder transformer with plain multi-head attention (plain MHA), which autoregressively predicts motion tokens conditioned on the input signal. We further compare it with a \textbf{causal MHA} variant to examine the effect of attention masking. Diffusion-Transformer (\textbf{DiT}) formulates motion generation as a denoising diffusion process with a transformer backbone. We also report \textbf{IPMG (w/o PetVAE)} to evaluate the contribution of the PetVAE module. The full model, \textbf{IPMG w/ MAA}, combines PetVAE with the proposed multi-modal motion attention mechanism. 
These variants allow us to isolate the effects of autoregressive modeling, PetVAE tokenization, and the proposed MAA mechanism.

\subsection{Quantitative Comparison}
\label{sec:quantitative}
\begin{table}[t]
\centering
\scriptsize
\caption{Quantitative comparison and ablation study of baseline architectures and design variants.}
\setlength{\tabcolsep}{4pt}
\renewcommand{\arraystretch}{1.15}
\begin{tabular}{lcccccccc}
\toprule
\multirow{2}{*}{\textbf{Architecture / Variant}} 
& \multirow{2}{*}{$FID_k\downarrow$} 
& \multirow{2}{*}{$FID_s\downarrow$} 
& \multicolumn{2}{c}{$R_{Prec.}^{Hand}\uparrow$}
& \multicolumn{2}{c}{$R_{Prec.}^{Body}\uparrow$}
& \multirow{2}{*}{$Div_k\uparrow$} 
& \multirow{2}{*}{$Div_s\uparrow$} \\
\cmidrule(lr){4-5}
\cmidrule(lr){6-7}
& & & Top-1 & Top-3 & Top-1 & Top-3 & & \\
\midrule
seq2seq-Transf. (plain MHA)
& 21.22 & 24.18 & 0.25 & 0.41 & 0.22 & 0.38 & 5.01 & 5.07 \\

Diffusion-Transf.                
& 64.37 & 67.94 & -- & 0.26 & -- & 0.23 & 4.42 & 4.50 \\

\midrule

IPMG w/ causal MHA
& 13.83 & 15.44 & 0.38 & 0.56 & 0.36 & 0.54 & \textbf{6.20} & 5.85 \\

IPMG w/o PetVAE   
& 14.15 & 16.21 & -- & 0.56 & -- & 0.52 & 5.62 & 5.70 \\

IPMG w/ MAA (Ours)
& \textbf{11.21} & \textbf{12.96} & \textbf{0.46} & \textbf{0.63} 
& \textbf{0.43} & \textbf{0.59} & 5.93 & \textbf{6.01} \\
\bottomrule
\end{tabular}
\label{tab:quant_compare}
\end{table}
Table~\ref{tab:quant_compare} presents the quantitative comparison with different baseline architectures. The proposed \textbf{IPMG} achieves the best performance across all evaluation metrics. Compared to the autoregressive \textbf{Seq2Seq-Transformer}, IPMG reduces $FID_k$ by 47.2\% (from 21.22 to 11.21) while significantly improving the alignment of motion conditions, increasing $R_{Prec.}^{hand}$ from 0.41 to 0.63, and $R_{Prec.}^{body}$ from 0.38 to 0.59.
The diversity of generated motions improves from 5.01 to 5.93, indicating that IPMG produces more varied, yet realistic, motion sequences.
Compared with the \textbf{causal MHA} variant, IPMG further improves $FID_k/FID_s$ from 13.83/15.44 to 11.21/12.96 and increases hand/body R-Precision from 0.56/0.54 to 0.63/0.59.
This indicates that allowing the model to access broader contextual information across the interaction sequence is beneficial for generating more realistic and condition-consistent pet motions.
The improvement is especially important for human-pet interaction modeling, where the pet's response often depends not only on preceding human actions but also on the overall interaction context.

\noindent Compared to the \textbf{DiT} baseline, which directly applies a generic diffusion backbone to motion generation, IPMG reduces $FID_k$ by 82.6\% and substantially improves alignment scores for hand and body motions (0.26 to 0.63, and 0.23 to 0.59).
This result suggests that generic diffusion architectures struggle to model structured human–pet interactions without appropriate motion priors.

\noindent Finally, removing the PetVAE module leads to consistent performance degradation across all metrics.
Adding PetVAE further reduces $FID_k$ by 20.8\% (14.15 to 11.21) while improving both alignment and diversity scores, demonstrating that the learned discrete motion representation helps stabilize training and improve motion realism.
\begin{table}[t]
\centering

\caption{Ablation study on input modalities. 
Using both body and hand inputs improves motion quality, alignment, and diversity compared to single-modality inputs.}
\setlength{\tabcolsep}{6pt}
\renewcommand{\arraystretch}{1.1}
\begin{tabular}{lcccccc}
\toprule
\textbf{Input} & $FID_k\downarrow$ & $FID_s\downarrow$ & $R_{Prec.}^{hand}@3\uparrow$ & $R_{Prec.}^{body}@3\uparrow$ & $Div_k\uparrow$ & $Div_s\uparrow$ \\
\hline
Body      & 13.48 & 15.72 & 0.44 & 0.57 & 5.36 & 5.41 \\
Hand      & 17.92 & 19.83 & 0.58 & 0.36 & 5.88 & 5.94 \\
Body+Hand & \textbf{11.21} & \textbf{12.96} & \textbf{0.63} & \textbf{0.59} & \textbf{5.93} & \textbf{6.01} \\
\toprule
\end{tabular}
\label{tab:input_ablation}
\end{table}
\subsection{Ablation Study}
\label{sec:ablation}
\noindent \textbf{Input modality ablation.}
\noindent Table~\ref{tab:input_ablation} presents an ablation study on different input modalities. Using both body and hand signals consistently yields the best performance across all metrics. When only body information is used, the model achieves better body-level alignment ($R_{Prec.}^{body}=0.57$) but performs worse on hand-level alignment ($R_{Prec.}^{hand}=0.44$), indicating that body signals alone provide global interaction context but lack the fine-grained cues required for modeling detailed human–pet interactions.


\noindent In contrast, using only hand input improves hand-level alignment ($R_{Prec.}^{hand}=0.58$) but significantly degrades body-level alignment ($R_{Prec.}^{body}=0.36$), indicating that local hand signals alone lack sufficient global motion context.

\begin{table}[t]
\centering
\scriptsize
\caption{Ablation on the PetVAE bone constraint. 
The full PetVAE improves motion quality and stability compared to the version without bone constraints.}
\setlength{\tabcolsep}{8pt}
\renewcommand{\arraystretch}{1.25}
\begin{tabular}{lcccc}
\hline
\textbf{} & $FID\downarrow$ & MPJPE & Vel. Err. \\
\hline
PetVAE (w/o bone)   & 12.53 & 7.26 & 1.79  \\
PetVAE         & 9.20 & 6.42 & 1.03 \\
\hline
\end{tabular}
\label{tab:vqvae_compare}
\vspace{-10pt}
\end{table}
\noindent \textbf{PetVAE ablation.} Table~\ref{tab:vqvae_compare} evaluates the effect of the bone constraint in PetVAE. Removing the bone constraint leads to degraded motion quality across all metrics, increasing FID from 9.20 to 12.53 and MPJPE from 6.42 to 7.26. In addition, the motion velocity error increases from 1.03 to 1.79, indicating less stable motion dynamics. These results show that incorporating bone constraints helps preserve kinematic consistency and improves the realism of the learned motion representation.

\begin{figure}[t]
    \centering
    \includegraphics[width=0.98\linewidth]{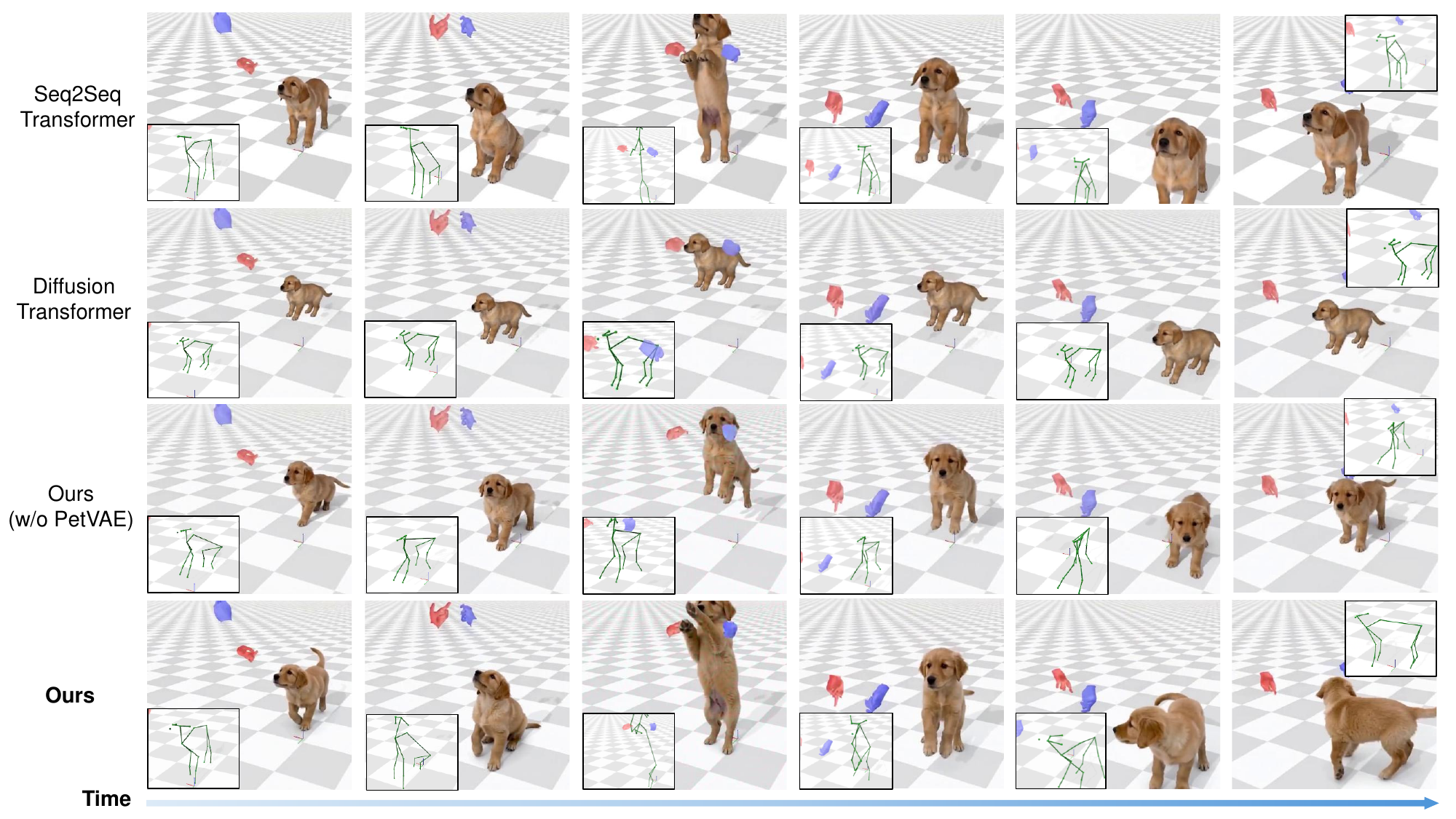}
    \caption{\textbf{Qualitative results.} We visualize InterPetMoGen outputs in a sequential command of "come->jump->turn" compared to other baselines. For each example, only the trainer's hands are rendered for better visibility of the dog.}
    \label{fig:qualitative}
\end{figure}
\subsection{Qualitative Results}
\label{sec:qualitative}
\noindent Figure~\ref{fig:qualitative} presents qualitative comparisons of generated dog motion sequences under a sequential command scenario (``come $\rightarrow$ jump $\rightarrow$ turn''). 
Our method produces the most natural and responsive behaviors among all compared approaches. 
The generated motions not only remain temporally smooth, but also clearly react to the trainer’s hand cues, including turning toward the command direction and executing the instructed actions.

\noindent In contrast, the Seq2Seq baseline can produce partially plausible motions, but shows limited responsiveness to input gestures. 
The dog’s orientation remains unchanged across the sequence, and actions such as turning are rarely observed.

\noindent Our model w/o PetVAE module is still able to somehow  react to the commands, but the resulting motions lack realism and appear less physically natural. 
Finally, the DiT baseline performs the worst among the compared methods, producing unstable and less meaningful motions, likely due to the limited amount of training data available for diffusion-based modeling in this setting.

\begin{table}[t]
\centering
\small
\setlength{\tabcolsep}{8pt}
\caption{\textbf{User study results.} Average participant ratings on a 7-point Likert scale evaluating the naturality, responsiveness, and overall quality of generated dog motions. Higher scores indicate better perceptual quality.}
\label{tab:userstudy}
\begin{tabular}{lccc}
\toprule
\textbf{Method} & \textbf{Naturality} $\uparrow$ & \textbf{Responsiveness} $\uparrow$ & \textbf{Overall} $\uparrow$ \\
\midrule
Seq2Seq-Transf. & 4.04 & 3.67 & 3.67 \\
DiT & 2.48 & 2.39 & 2.12 \\
IPMG (w/o PetVAE) & 3.82 & 3.64 & 3.70 \\
IPMG (Ours)  & \textbf{6.58} & \textbf{6.55} & \textbf{6.63} \\
\bottomrule
\end{tabular}
\end{table}

\subsection{User Study}
\label{sec:userstudy}

\noindent Because pet behavior does not deterministically follow human gestures, evaluating the perceptual quality of generated motions is essential. 
We therefore conducted a user study with 12 participants to assess the naturalness and appropriateness of the generated dog responses.

\noindent Each participant evaluated three randomly selected human-interaction input sequences and the corresponding dog motions generated by different methods. To make the evaluation more intuitive, we rendered the skeletal motions into realistic dog videos using a finetuned video DiT model (Figure~\ref{fig:qualitative}). Participants rated each result on a 7-point Likert scale for \emph{naturality}, \emph{responsiveness}, and \emph{overall motion quality}.

\noindent The results are summarized in Table~\ref{tab:userstudy}.
Our full model achieves the highest scores across all criteria, with 6.58 in naturality, 6.55 in responsiveness, and 6.63 in overall quality, clearly outperforming the strongest baseline, Seq2Seq-Transformer (4.04, 3.67, and 3.67).
This suggests that explicitly modeling human-pet interactions with our proposed components leads to more natural and responsive pet motions than directly applying generic seq-to-seq motion generation models.
Removing PetVAE also degrades performance across all criteria, confirming its importance for capturing realistic and diverse pet motion patterns during human-pet interaction.
A repeated-measures ANOVA further shows significant condition effects for naturality ($F(3,6)=5.78$, $p=.033$), responsiveness ($F(3,6)=24.91$, $p<.001$), and overall quality ($F(3,6)=9.39$, $p=.011$), indicating that the observed differences among methods are statistically significant.

\noindent Importantly, these subjective results are consistent with our quantitative evaluation (Section~\ref{sec:experiments}) and the qualitative comparisons in Figure~\ref{fig:qualitative}. 
Together, these results demonstrate that IPMG not only improves objective metrics but also produces perceptually more natural and responsive dog motions in human–pet interaction scenarios.




\section{Limitations and Future Work}
\label{sec:limitations}
While InterPet4D is the first dataset of its kind, it has several limitations.

\noindent \textbf{First}, the dataset currently covers only dogs; extending to other pet species (\eg, cats) would increase generality but requires species-specific pose models.
\textbf{Second}, the current framework does not model physical contact forces between human and dog, which are important in petting and playing interactions. Incorporating physics-based constraints~\cite{rueegg2023bite} could improve physical plausibility.
\textbf{Third}, our model generates motion clips of fixed length (10 seconds), extending to variable-length or autoregressive generation would enable modeling of longer interactions.

\section{Conclusion}
\label{sec:conclusion}

We presented InterPet4D, the first large-scale multimodal 4D dataset for human--pet interaction, featuring synchronized 12-view third-person and egocentric RGB video from Ray-Ban 2 glasses, 3D human body and hand motion, 3D dog pose, and audio across 23 participants and 13 dogs. 
We established a systematic interaction taxonomy and annotation pipeline, providing a standardized benchmark for human-pet interaction research. 
We further introduced InterPetMoGen, an autoregressive model that leverages discrete motion representations to synthesize plausible pet responses conditioned on human motion and audio. 
We hope InterPet4D will facilitate future research on multimodal human-pet interaction modeling.


\section{Acknowledgment}
This work was supported in part by JST ASPIRE JPMJAP2404, JST CRONOS JPMJCS24N8, Crescent Inc., and the Shanda Interactive Intelligence Collaborative Research Cluster. This study was carried out using the TSUBAME4.0 supercomputer at Institute of Science Tokyo.


\bibliographystyle{splncs04}
\bibliography{main}
\newpage
\appendix

This supplementary material contains the following sections:
\begin{itemize}
    \item \textbf{A.} Network Architecture \& Training Details
    \item \textbf{B.} Dog Skeleton Definition
    \item \textbf{C.} Multi-GoPro Capture \& Synchronization Details
    \item \textbf{D.} Dog Shape \& Motion Analysis
    \item \textbf{E.} Audio Modality Ablation
    \item \textbf{F.} Dog Motion FID Model
    \item \textbf{G.} Ethical Statement
    \item \textbf{H.} Additional Qualitative Results (Supplementary Video)
\end{itemize}

\section{Network Architecture \& Training Details}
\label{sec:supp_architecture}

We provide detailed architecture specifications and training hyperparameters for each component of InterPetMoGen.

\paragraph{PetVAE.}
The PetVAE encoder takes a pet motion sequence of shape $(T, 60)$ (\ie, 20 keypoints $\times$ 3 coordinates) as input. We adopt an encoder-decoder architecture backbone with a single-level EMA vector quantization bottleneck. The encoder and decoder each consist of 4 stages with channel multipliers $[1, 2, 4, 4]$ and a base dimension of 128, yielding a latent dimension of 128 with a codebook size of 512. Temporal downsampling (factor $4\times$) is applied in the first two encoder stages via strided convolutions, and symmetrically upsampled in the decoder. Each stage contains 2 residual blocks with RMSNorm and SiLU activations. 

\paragraph{Human Motion Tokenizer.}
Human body and hand motions are tokenized independently using two separate VQ-VAE modules sharing the same architecture but with different input dimensions.
The body tokenizer takes SMPL joint rotations and positions $(\theta^{body}, \mathbf{J}^{body})$ as input, where $\theta^{body} \in \mathbb{R}^{T \times D_b}$ and $\mathbf{J}^{body} \in \mathbb{R}^{T \times J_b \times 3}$.
The hand tokenizer takes MANO joint rotations and positions,
denoted as $\mathbf{x}^{hand} = (\theta^{hand}, \mathbf{J}^{hand})$, where $\theta^{hand} \in \mathbb{R}^{T \times D_h}$ and $\mathbf{J}^{hand} \in \mathbb{R}^{T \times J_h \times 3}$.
Both tokenizers use $4\times$ temporal downsampling and a codebook of size 512. Each module includes two decoder heads: one for joint rotations and one for joint positions.

\paragraph{Translation Tokenizer.}
A lightweight VQ-VAE encodes the global root translation sequence $\mathbf{T} \in \mathbb{R}^{T \times 3}$ into discrete tokens. This module uses $4\times$ temporal downsampling to capture coarse motion dynamics and a codebook of size 512.

\paragraph{Audio Tokenizer.}
Audio features are extracted using the pretrained MERT~\cite{li2023mert} model, yielding frame-level features of dimension $d_a = 1024$. These features are projected into token embeddings of dimension 512 via a linear layer and fed to the transformer through cross-attention.

\paragraph{Autoregressive Transformer (GPT).}
The prefix-LM GPT model has 30.3M parameters and consists of 6 transformer blocks. Each block contains a self-attention layer with 8 heads, a cross-attention layer for audio conditioning, and a feed-forward network with hidden dimension 2048. The input sequence consists of 375 tokens: 150 hand tokens (75 left + 75 right), 75 SMPL body tokens, 75 pet-human relative translation tokens, and 75 pet motion tokens. Human motion tokens (hand + body) use bidirectional attention as global conditioning (prefix), while relative translation and pet tokens are generated autoregressively with causal masking. During inference, we apply top-$k$ sampling with $k=50$ and temperature $\tau=1.0$.

\paragraph{Training Details.}
All models are implemented in PyTorch and trained on a single NVIDIA H100 GPU.
\begin{itemize}[leftmargin=15pt,itemsep=1pt]
    \item \textbf{VQ-VAEs (Stage 1):} Trained for 500-1000 epochs using the Adam optimizer with a learning rate of $3 \times 10^{-5}$, $\beta_1 = 0.5$, $\beta_2 = 0.999$, weight decay of 0.01, and batch size 128. A MultiStepLR schedule reduces the learning rate by half at epochs 100, 200, and 300. The commitment loss weight $\beta = 0.02$ is linearly warmed up over the first 50 epochs.
    \item \textbf{GPT (Stage 2):} Trained for 300 epochs using the AdamW optimizer with a learning rate of $3 \times 10^{-4}$, batch size 32, weight decay 0.01, and gradient clipping with max norm 1.0. A MultiStepLR schedule reduces the learning rate by half at epochs 100 and 200.
\end{itemize}

\section{Dog Skeleton Definition}
\label{sec:supp_skeleton}

\begin{figure}[t]
    \centering
    \includegraphics[width=0.6\linewidth]{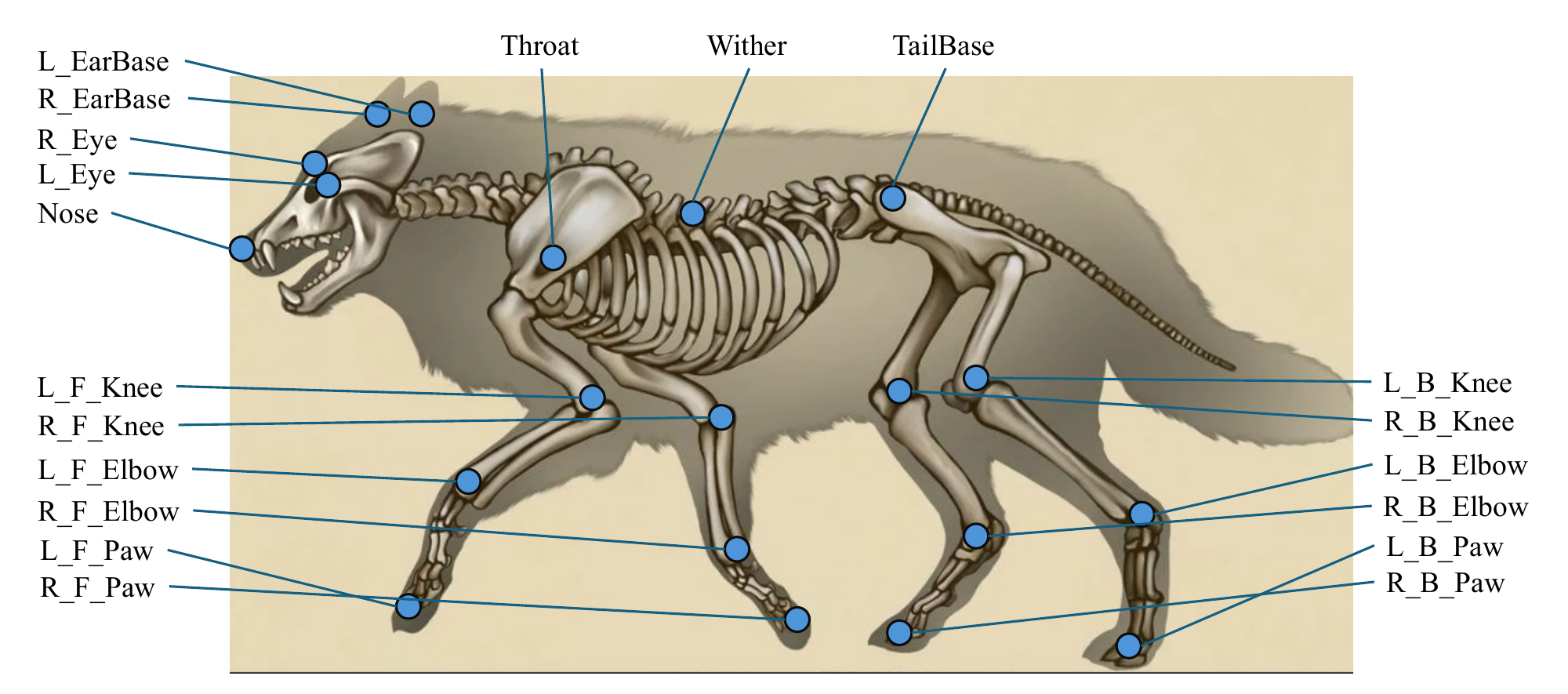}
    \caption{\textbf{Dog skeleton topology.} Our 20-keypoint skeleton covers the facial region, torso, limbs, and tail base.}
    \label{fig:skeleton}
\end{figure}

We adopt a 20-keypoint dog skeleton based on the AnimalPose definition, as illustrated in Figure~\ref{fig:skeleton}. The keypoints correspond to major anatomical landmarks of the dog, including the facial region, torso, limbs, and tail base. Table~\ref{tab:skeleton} lists all keypoints together with their parent keypoints, defining the kinematic hierarchy. We treat the \textit{Withers} joint as the root of the skeleton, and construct the skeletal graph by connecting each keypoint to its parent according to the kinematic tree. We also provide a \texttt{data\_sample} file in the supplementary material for reference.

\begin{table}[t]
\centering
\small
\caption{\textbf{Dog skeleton definition.} The 20 AnimalPose~\cite{cao2019cross} keypoints and their parent connections following the AnimalPose skeleton. (We follow the it's definition, which does not include tail tip keypoints.)}
\label{tab:skeleton}
\setlength{\tabcolsep}{4pt}
\begin{tabular}{clc|clc}
\toprule
\textbf{Idx} & \textbf{Keypoint} & \textbf{Parent} & \textbf{Idx} & \textbf{Keypoint} & \textbf{Parent} \\
\midrule
0  & L\_Eye        & - & 10 & L\_B\_Knee   & 6   \\
1  & R\_Eye        & 0  & 11 & R\_B\_Knee   & 6  \\
2  & L\_EarBase    & 0  & 12 & L\_F\_Elbow  & 8  \\
3  & R\_EarBase    & 1  & 13 & R\_F\_Elbow  & 9  \\
4  & Nose          & 2  & 14 & L\_B\_Elbow  & 10 \\
5  & Throat        & 0  & 15 & R\_B\_Elbow  & 11 \\
6  & TailBase      & 7  & 16 & L\_F\_Paw    & 12 \\
7  & Withers       & 5  & 17 & R\_F\_Paw    & 13 \\
8  & L\_F\_Knee    & 5  & 18 & L\_B\_Paw    & 14 \\
9  & R\_F\_Knee    & 5  & 19 & R\_B\_Paw    & 15 \\
\bottomrule

\end{tabular}
\end{table}

\section{Multi-GoPro Capture \& Synchronization Details}
\label{sec:supp_sync}

\paragraph{Camera Calibration.}
The 12 GoPro cameras were calibrated using a checkerboard pattern with a ($7\times4$) grid and a cell size of 0.1. We first detected chessboard corners in the synchronized calibration images, then estimated the intrinsic parameters of each camera, including lens distortion, and solved for the extrinsic parameters in a common coordinate system. The average calibration error reported across the 12 cameras was 0.245 pixels (approximately 0.25 pixels) in 1080p resolution.

\paragraph{Camera Synchronization}
Our capture system consists of multiple GoPro cameras controlled via the OpenGoPro HTTP API.
Prior to each recording session, we synchronize the internal clocks of all cameras to the host PC's system time through the API's \texttt{set\_date\_time} endpoint, ensuring a consistent time reference across devices. Recording is then triggered simultaneously on all cameras using parallel HTTP requests, so that each camera begins capturing at approximately the same moment. Each GoPro embeds a timecode in the recorded video stream metadata, reflecting its synchronized internal clock. In post-processing, we extract these embedded timecodes using \texttt{ffprobe} and compute per camera temporal offsets relative to the latest-starting camera. All video streams are then front-trimmed accordingly and cropped to their common overlapping duration, yielding frame-level synchronized multi-view footage. For the egocentric camera, synchronization with the multi-view system is achieved manually using a clapping gesture recorded at the beginning of each session as a visual cue.

\section{Dog Shape \& Motion Analysis}
\label{sec:supp_shape_motion}

\begin{figure}[t]
    \centering
    \begin{subfigure}[t]{\linewidth}
      \centering
      \includegraphics[width=\linewidth]{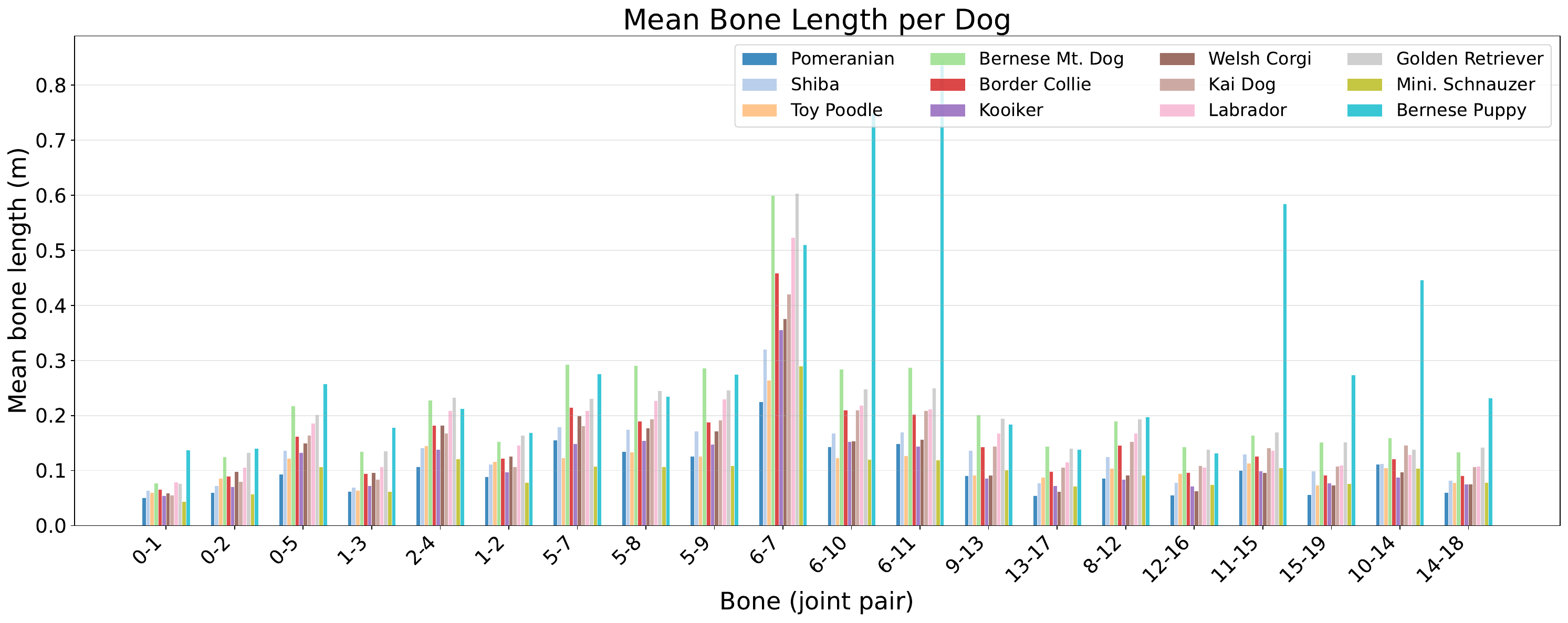}
      \caption{Mean bone length per breed.}
      \label{fig:bone_bar}
    \end{subfigure}
    \\[4pt]
    \begin{subfigure}[t]{0.48\linewidth}
      \centering
      \includegraphics[width=\linewidth]{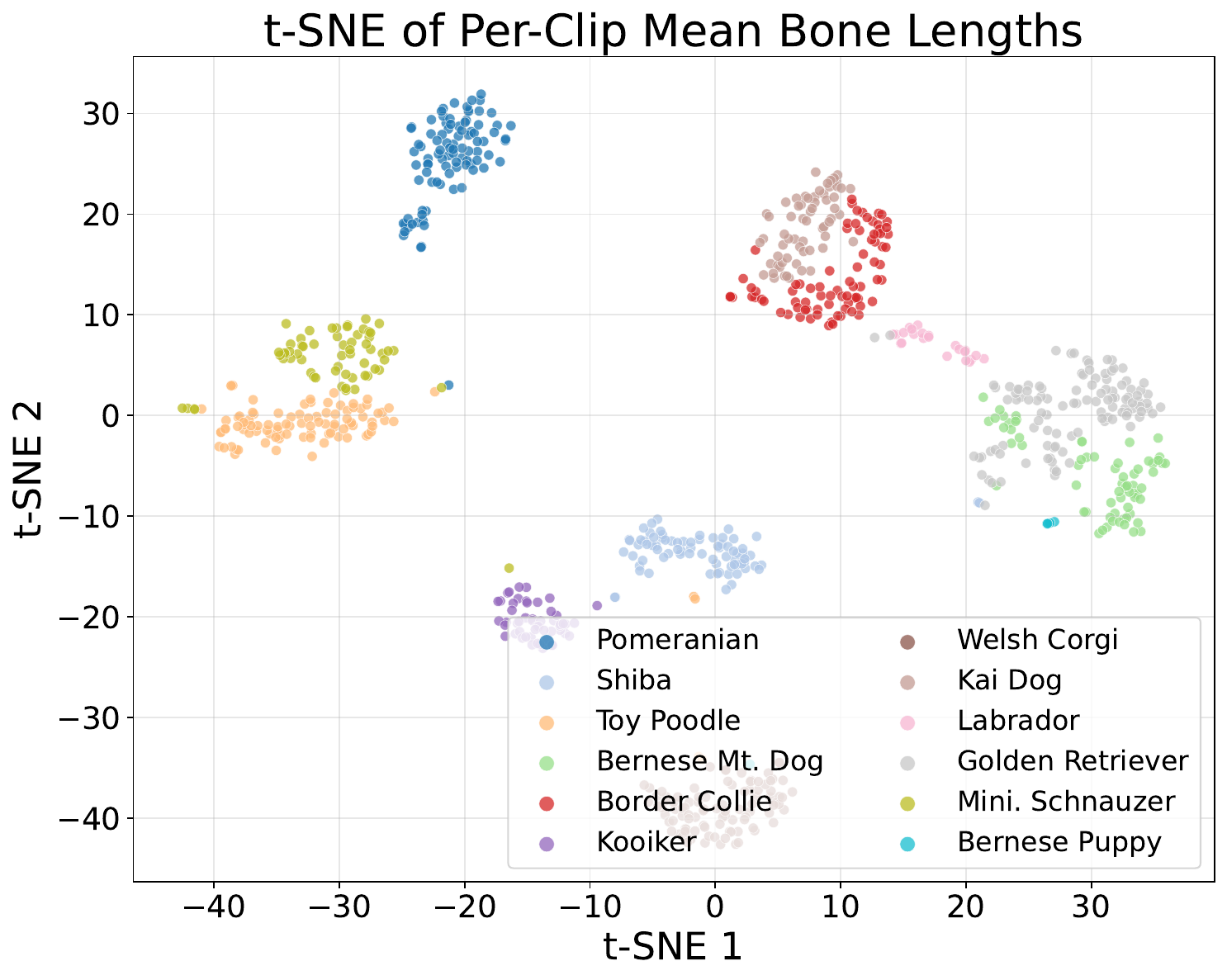}
      \caption{t-SNE of per clip bone lengths.}
      \label{fig:bone_tsne}
    \end{subfigure}
    \hfill
    \begin{subfigure}[t]{0.48\linewidth}
      \centering
      \includegraphics[width=\linewidth]{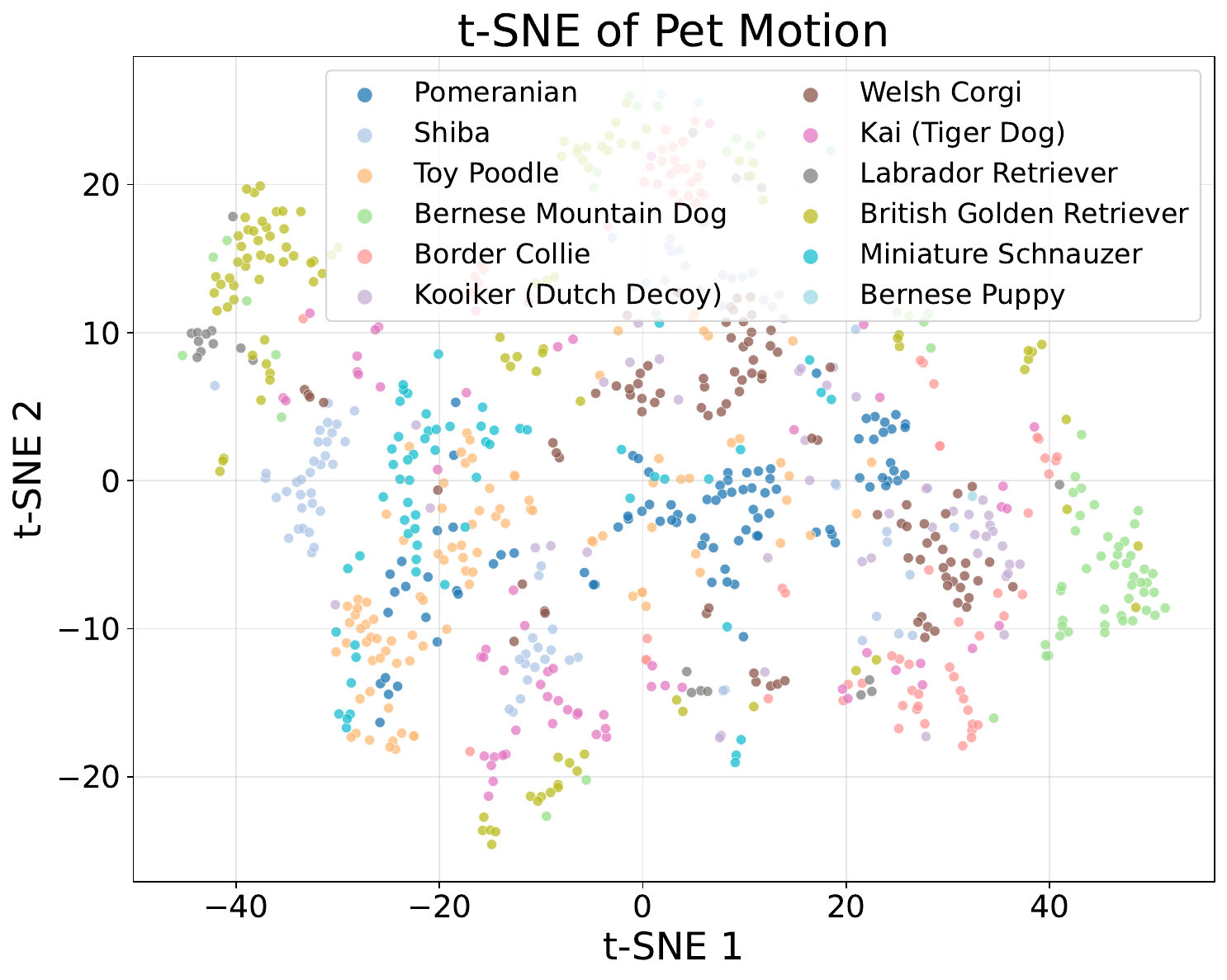}
      \caption{t-SNE of per clip motion features.}
      \label{fig:motion_tsne}
    \end{subfigure}
    \caption{\textbf{Shape and motion analysis.} (a) Mean bone lengths differ markedly across the 12 breeds, capturing diverse body proportions. (b) per clip bone length vectors form tight per dog clusters in t-SNE space, demonstrating consistent and accurate shape estimation. (c) Motion features are broadly distributed and overlapping across breeds, reflecting the rich motion diversity in our dataset.}
    \label{fig:shape_motion}
\end{figure}

Our dataset captures 12 dogs spanning a wide range of breeds, from small breeds such as Pomeranian and Toy Poodle to large breeds such as Bernese Mountain Dog and Labrador Retriever. Note that we exclude one puppy from the original 13 dogs due to its limited number of recorded sequences, and conduct the following analysis on the remaining 12. We analyze the shape and motion characteristics of the captured data to validate the quality and diversity of our dataset.

\paragraph{Shape consistency.}
For each dog, we compute the mean bone length of all 20 skeleton bones across every clip.
Figure~\ref{fig:bone_bar} reports the per bone mean lengths grouped by breed. The bone-length profiles vary substantially across breeds, reflecting genuine differences in body proportions, for example, limb bones (6-7, 6-10, 6-11) of the Bernese Mountain Dog are roughly three times longer than those of the Pomeranian. To further assess shape estimation consistency, we apply t-SNE to the 20-dimensional per clip mean bone-length vectors (Figure~\ref{fig:bone_tsne}). Clips from the same dog form tight, well separated clusters, confirming that our skeleton estimation pipeline produces consistent body shapes within each individual and clearly distinguishes the 12 distinct body types.

\paragraph{Motion diversity.}
We extract motion features for each clip, including the mean pose, pose variability, and mean velocity, and project them using t-SNE (Figure~\ref{fig:motion_tsne}).
In contrast to the shape embeddings, the motion embeddings exhibit broadly distributed and partially overlapping clusters across breeds. This suggests that while individual dogs display characteristic motion tendencies that form loosely grouped regions in the embedding space, the overall motion distribution remains diverse, spanning a wide range of behaviors across the dataset.

\section{Audio Modality Ablation}
\label{sec:supp_audio_ablation}
To evaluate the contribution of audio conditioning, we compare the full IPMG model (body + hand + audio) with a variant that removes the audio input. As shown in Table~\ref{tab:audio_ablation}, the absence of audio results in consistent degradation across motion quality, alignment, and diversity metrics. The increase in $FID_k$ and the decrease in $R_{Prec.}^{hand}$ indicate that audio provides complementary temporal cues that help align human gestures with corresponding pet responses. This is particularly relevant in voice-driven interactions such as calling or commanding.

\begin{table}[t]
\centering
\small
\caption{\textbf{Audio modality ablation.} Removing audio conditioning leads to consistent degradation in motion quality and alignment, suggesting that audio provides complementary temporal cues for modeling human-pet interactions.}
\setlength{\tabcolsep}{5pt}
\begin{tabular}{lcccccc}
\toprule
\textbf{Input} & $FID_k\downarrow$ & $FID_s\downarrow$ & $R_{Prec.}^{hand}\uparrow$ & $R_{Prec.}^{body}\uparrow$ & $Div_k\uparrow$ & $Div_s\uparrow$ \\
\midrule
IPMG (w/o Audio) & 13.69 & 14.24 & 0.60 & 0.58 & 5.73 & 5.81 \\
IPMG (Full) & \textbf{11.21} & \textbf{12.96} & \textbf{0.63} & \textbf{0.59} & \textbf{5.93} & \textbf{6.01} \\
\bottomrule
\end{tabular}
\label{tab:audio_ablation}
\end{table}

\section{Dog Motion FID Model}
\label{sec:supp_fid}

Following the standard practice of using a pretrained classifier as a feature extractor for Fr\'{e}chet Inception Distance (FID)~\cite{heusel2018ganstrainedtimescaleupdate}, we train a dog identity classifier on our motion dataset to serve as the backbone for motion FID computation. The classifier takes normalized motion sequences of shape $(T, 60)$ (20 keypoints $\times$ 3 coordinates) as input, and processes them through a 1D convolutional ResNet encoder (4 residual blocks, 512 channels) with temporal downsampling. Global average pooling is applied to the encoded features to obtain a 512-dimensional representation, which is then mapped to 12 dog identity classes via a linear head, trained with cross entropy loss. After training, we discard the classification head and use the penultimate 512-dimensional features to compute FID between the generated and ground-truth motion distributions, measuring both the quality and diversity of the generated motions.

\section{Ethical Statement}
\label{sec:supp_ethics}

\paragraph{Human participants.}
All human participants provided written informed consent prior to data collection. The study protocol was reviewed and approved by the local Institutional Review Board. Participants were informed of the data capture procedure, the intended use of the data for research purposes, and their right to withdraw at any time. 

\paragraph{Animal welfare.}
All dog participants were accompanied by their owners or trained handlers throughout the recording sessions. No aversive training methods or restraints were used. Dogs were free to move voluntarily within the capture space and were given breaks between sessions. The interaction protocol was designed in consultation with professional dog trainers to ensure that all tasks fall within the range of typical obedience exercises. 

\paragraph{Data release.}
The dataset will be released under a license for non-commercial research purposes. Users must agree to a data use agreement prohibiting redistribution and use for surveillance or biometric identification.

\section{Additional Qualitative Results (Supplementary Video)}
\label{sec:supp_qualitative}
Additional qualitative results are provided in the supplementary video \\(\texttt{interpet4d\_supp\_video\_final.mp4}).

\clearpage

\end{document}